\documentclass[10pt, a4paper]{article}

\usepackage[]{lrec-coling2024}
\usepackage{multicol}
\usepackage{multirow}
\usepackage{lipsum}
\usepackage{graphicx}
\usepackage{pifont}
\usepackage{booktabs}
\usepackage{amsmath}
\usepackage{amssymb}

\title{BanglaAutoKG: Automatic Bangla Knowledge Graph Construction with Semantic Neural Graph Filtering}

\name{Azmine Toushik Wasi$^1$, Taki Hasan Rafi$^2$, Raima Islam$^3$, Dong-Kyu Chae\thanks{*Corresponding author}$^{2*}$} 

\address{$^1$Shahjalal University of Science and Technology, Bangladesh\\ $^2$Hanyang University, Republic of Korea $^3$BRAC University, Bangladesh \\
         azmine32@student.sust.edu, \{takihr, dongkyu\}@hanyang.ac.kr,\\
         raima.islam@g.bracu.ac.bd\\
         }

\abstract{
Knowledge Graphs (KGs) have proven essential in information processing and reasoning applications because they link related entities and give context-rich information, supporting efficient information retrieval and knowledge discovery; presenting information flow in a very effective manner. Despite being widely used globally, Bangla is relatively underrepresented in KGs due to a lack of comprehensive datasets, encoders, NER (named entity recognition) models, POS (part-of-speech) taggers, and lemmatizers, hindering efficient information processing and reasoning applications in the language. Addressing the KG scarcity in Bengali, we propose BanglaAutoKG, a pioneering framework that is able to automatically construct Bengali KGs from any Bangla text. We utilize multilingual LLMs to understand various languages and correlate entities and relations universally. By employing a translation dictionary to identify English equivalents and extracting word features from pre-trained BERT models, we construct the foundational KG. To reduce noise and align word embeddings with our goal, we employ graph-based polynomial filters. Lastly, we implement a GNN-based semantic filter, which elevates contextual understanding and trims unnecessary edges, culminating in the formation of the definitive KG. Empirical findings and case studies demonstrate the universal effectiveness of our model, capable of autonomously constructing semantically enriched KGs from any text. Data and code are available here: \url{https://github.com/azminewasi/BanglaAutoKG}.
 \\ \newline  }

\begin{document}

\maketitleabstract
\vspace{-3mm}
\section{Introduction}
Knowledge Graphs (KGs) are semantic graphs consisting of large collections of factual entities and relations, which depict knowledge of real-world objects. 
KGs provide well-organized human knowledge for applications like search engines \citep{xiong2017explicit}, recommendation systems \citep{wang2018ripplenet, sakg}, and question answering \citep{bao2016constraint}. KGs have been instrumental in enabling efficient information retrieval and knowledge discovery by connecting related entities and providing context-rich information by covering domains like entity typing \citep{xu2018neural, ren2016label}, entity linking \citep{ganea2017deep,le2018improving} and relation extraction \citep{zeng2015distant, zhou2016attention}.

Automatic KG generation is important because it reduces the need for large labeled datasets, enables transfer learning, and provides explanations \citep{eswtGEWSAGREWAG}. Currently, KG construction involves manual effort and is time-consuming, hindering its application in certain situations. Automating this process would benefit small organizations and improve the state of the art in constructing KGs from text \citep{khorashadizadeh2023exploring}. 
Automatic KG generation that utilizes probabilistic methods such as RIBE \citep{peng-etal-2019-improving} or employs neural networks based on embeddings for e.g. NetTaxo \citep{shang2020nettaxo} and SSE \citep{guo2015semantically} are such benchmark works done for identifying entity relationships. 

\begin{figure*}[t] 
\includegraphics[scale=0.58]{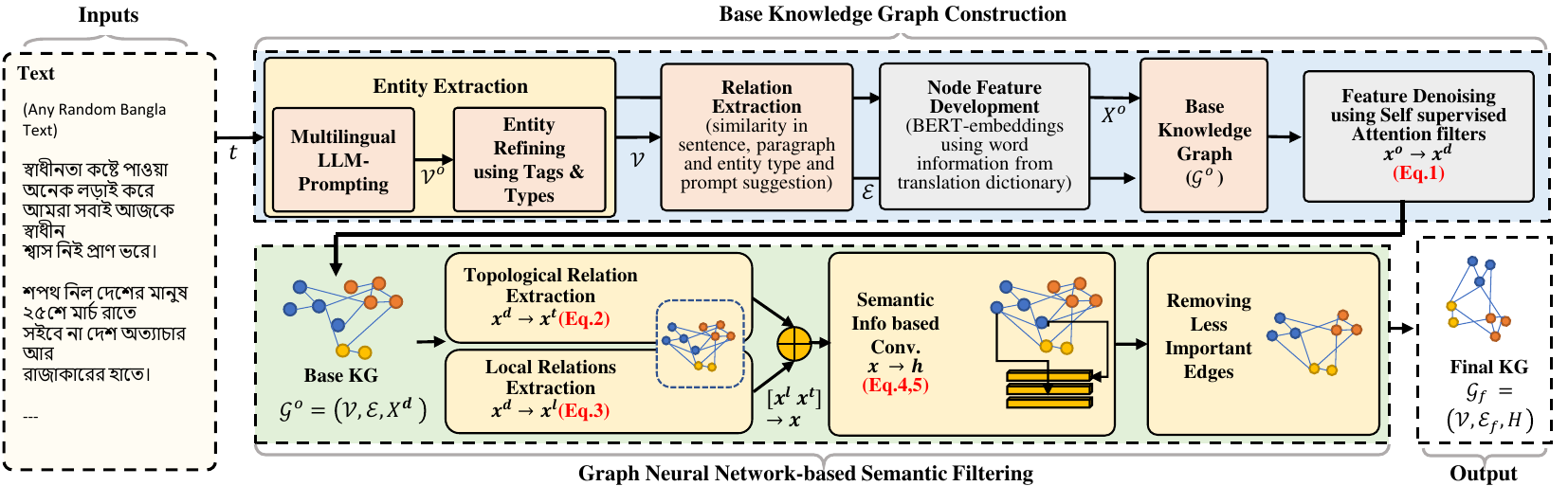} \label{fig:main}
\vspace{-3mm}
\caption{The overall framework of our BanglaAutoKG. It involves passing text data through a multilingual LLMs to obtain entities and entity types, which are used to build a base KG with dictionary-based BERT embeddings. This graph is then semantically filtered using local neighborhood and topological relations to extract important nodes and edges, resulting in the final KG.}
\end{figure*}

While KGs have seen a lot of advancement in the field of NLP, it is yet to gain traction in Bangla due to its limited resources. It is the sixth most spoken language in the world, with almost 300 million speakers. Though there are some prominent works on Bangla language like BanglaBERT \citep{bhattacharjee-etal-2022-banglabert}, BanglaT5 and BanglaNLG \citep{bhattacharjee2023banglanlg}; but they cannot be applied universally, as the datasets used such as SentNoB \citep{islam2021sentnob}, a discourse mode detection dataset \citep{sazzed-2022-annotated} and BanglaRQA \citep{ekram2022banglarqa} are limited and not universal, comparing to the task required.
Connecting everything, lack of comprehensive resources in the Bengali language, including a Bangla KG dataset and a methodology for KG development, is compounded by the absence of a universal dataset. Such a universal dataset should encompass a wide array of textual contexts, such as blogs, news, Wikipedia, poems, stories, personal writings, and memoirs for universality. Moreover, the current limitations in encoders and NER (Named Entity Recognition) models pose challenges in encoding diverse text effectively and recognizing entities and relations accurately, particularly in the presence of novel names, places, and contexts not adequately represented in their training data; which is crucial for developing universal KGs automatically.

To solve the above mentioned problems and develop a universal automatic KG generation method for Bangla, we propose \textbf{BanglaAutoKG}.
Our novel approach leverages the capabilities of multilingual Large Language Models (LLMs) to extract entities and their relationships from diverse textual sources, underscoring the model's universality. We complement this by employing translation dictionaries to identify semantic similarities and distinctions between words, which aids in the construction of foundational relations for our KG. To establish word embeddings, we employ a pre-trained BERT model. However, these embeddings may exhibit noise due to the model's lack of context awareness.
To mitigate this noise, we adopt a Graph Neural Network (GNN)-based feature denoising method, utilizing self-supervised attention filtering. Subsequently, we design a GNN-based semantic filtering technique to identify and remove less significant connections, refining the initial KG into a more semantically enriched representation. 

It is important to note that this entire process is fully automated and operates independently of human intervention. The universality of our model is a notable strength, as it can proficiently process and extract insights from a wide array of textual content. 

Our contributions are summarized below: 
\begin{itemize}
  \setlength{\itemsep}{1pt}
  \setlength{\parskip}{0pt}
  \setlength{\parsep}{0pt}
    \item We are the first to develop a novel and universal automatic KG generation framework with semantic-filtered for Bangla language, which can be effortlessly utilized for texts of any length and context.
    \item We construct a universal KG by utilizing multilingual LLMs for entity extraction, dictionary-based relation building, pre-trained BERT based feature development and feature alignment within entities through GNN-based feature denoising. We also develop a semantic filtering method to streamline KGs by removing unnecessary edges.
    \item Our experiments show that BanglaAutoKG is able to construct Bengali KGs from text automatically and very effectively.
\end{itemize}

\vspace{-3mm}
\section{Approach}

Figure \ref{fig:main} illustrates the overview of our BanglaAutoKG. We split the process into two parts: KG construction and semantic filtering. 
\vspace{-1mm}
\subsection{Knowledge Graph Construction}
\noindent \textbf{Entity and Relation Extraction.}
Given a paragraph, article or any text $t$, we process it using a multilingual LLM (ChatGPT or Bard) to extract different entities ($\mathcal{V}^o$) from the text to get entities and relations. Afterward, we refine the entity set ($\mathcal{V}^o$) using tags and types from the LLM and receive $\mathcal{V}$, by removing certain part of speeches, removing very long entities, etc.
We then establish initial connections set $\mathcal{E}$ based on the similarity between sentences, paragraphs, entity types and LLM relation suggestions. 

\bigskip
\noindent \textbf{Node Features Development.}
For each entity, we utilize a Bangla-to-English translation dictionary\footnote{\url{https://github.com/MinhasKamal/BengaliDictionary}} to uncover meaning and related words. Using this information, we leverage a pre-trained BERT \cite{DBLP:conf/naacl/DevlinCLT19} model to generate feature vectors for these entities, creating an initial feature matrix $X^o$. 

\bigskip
\noindent \textbf{Base Knowledge Graph.}
As a result, $\mathcal{V}$ denotes the ensemble of nodes (entities) $\{v_1, v_2, v_3, \cdots v_N\}$, and $\mathcal{E}$ designates the collection of edges (relationships among entities) $\{e_1, e_2, e_3, \cdots e_M\}$, where $N$ and $M$ denote the number of nodes and edges, respectively. $\mathcal{V}$, $\mathcal{E}$, and $X^o$ form our base KG $\mathcal{G}^o = (\mathcal{V}, \mathcal{E}, X^o)$, and its adjacency matrix is $\mathcal{A}$ where an element $\mathcal{A}_{ij}=1$ if there exists an edge connecting $v_i$ and $v_j$.

\bigskip
\noindent \textbf{Feature Denoising.}
Given that these feature vectors lack context, as they are derived solely from individual words and their synonyms rather than initial text $t$, they tend to be noisy and less effective. To mitigate this issue, we apply a self-supervised graph attention filter \citep{kim2021howwfwf} to $\mathcal{G}^o$, implementing a graph feature denoising technique. The GNN-based feature denoiser uses negative edge sampling with a cross-entropy based self-supervision loss to enhance the node features. For each node $v_i$, the feature vector $x^o_i$ undergoes a self-supervised transformation, resulting in $x^d_i$. The equation for the process is:
\begin{equation}\label{eq:feature-denoising}
\begin{aligned}
{x^d_i} =\alpha_{ii} W_{FD} x^o_i+\sum_{j \in \mathcal{N}(i)} \alpha_{ij} W_{FD} {x^o_j} 
\quad\quad\quad\quad
\\ \alpha_{ij} =\frac{\exp \left(\operatorname{Leaky} \operatorname{ReLU}\left(e_{ij}\right)\right)}{\sum_{k \in \mathcal{N}(i) \cup\{i\}} \exp \left(\operatorname{LeakyReLU}\left(e_{ik}\right)\right)} 
\quad\quad\quad
\\ e_{ij} = {\mathcal{A}}^{\top}\left[W_{FD} x_i \| W_{FD} {x_j}\right] \cdot \sigma\left(\left(W_{FD} x_i\right)^{\top}\cdot W_{FD} {x_j}\right) 
\end{aligned}
\end{equation}
where $W_{FD}$ denotes the model parameters of this layer and $\mathcal{N}(i)$ indicates the set of neighbors of $v_i$. This feature denoising process enriches the connectivity and resilience of the features, yielding the refined graph representation. As a result, we obtain a base KG with aligned features, symbolized as $\mathcal{G} = (\mathcal{V}, \mathcal{E}, X^d)$, where $X^d \in \mathbb{R}^{N \times F}$ is the collective node feature matrix for the entire graph and $F$ is the length of a feature vector. 




\subsection{Semantic Filtering}
In order to remove low-quality edges, we design a semantic graph filtering method that re-weights the edges Using topological and local neighborhood information with an attention-based convolution.

\bigskip
\noindent \textbf{Topological Relation Extraction.}
To implement semantic filtering, our initial step involves the extraction of topological relations using an attention-based convolution approach. For each node 0 $v_i$ within the foundational graph $\mathcal{G}$, the feature vector $x^d_i$ undergoes a transformation through a graph convolutional operator utilizing higher order features \cite{10.1609/aaai.v33i01.33014602}, resulting in $x^t_i$. This can be performed by:
\begin{equation} \label{eq:topological-relations}
    {x^t_i}=W_{TR}^{\top} \sum_{j \in \mathcal{N}(i) \cup\{i\}} \frac{e_{j, i}}{\sqrt{\hat{d}_j \hat{d}_i}} {x}^d_j\\
\end{equation}
where $\hat{d}_i=1+\sum_{j \in \mathcal{N}(i)} e_{j, i}$ and $e_{j, i}$ denotes the edge weight (in our case, it is $1$); $W_{TR}$ is a model parameter for this transformation layer.

\bigskip
\noindent \textbf{Local Relation Extraction.}
We further extract local neighborhood relations to enhance the construction of a more comprehensive KG. Similar to the previous procedure, for each node $v_i$, the feature vector $x^d_i$ undergoes a transformation using local neighborhood relations using spectral filtering \cite{cnn_graph}, resulting in $x^l_i$.  
This transformation is expressed by:
\begin{equation} \label{eq:local-neighbourhood-relations}
{X}^l=\sum_{k=1}^K {Z}^{(k)} \cdot W_{NR}^{(k)}
\end{equation}
Here, ${Z}^{(k)}$ is computed recursively by ${Z}^{(1)}={X^d}$, ${Z}^{(2)}=\hat{{L}} \cdot {X^d}$, $...$, ${Z}^{(k)}=2 \cdot \hat{{L}} \cdot {Z}^{(k-1)}-{Z}^{(k-2)}$ where $\hat{{L}}$ denotes the scaled and normalized Laplacian $\frac{2 {L}}{\lambda_{\max }}-{I}$. $X^l$ is the transformed feature matrix ($X^l = \{x^l_1, x^l_2,...,x^l_N\}$) and $W_{NR}$ is a model parameter for this process. $k$ is a hyperparameter denoting the filter size. 

\bigskip
\noindent \textbf{Semantic Information Convolution.}
The combination of $X^t$ ($X^t = \{x^t_1, x^t_2,...,x^t_N\}$, each obtained from Eq. (2)) and $X^l$ results in a unified feature representation, denoted as $X \in \mathbb{R}^{N \times 2F}$ ($X=[X^t, X^l]$). Subsequently, an attention-based \cite{veličković2018graph} semantic information convolution is applied to obtain the final node feature $h\in \mathbb{R}^{F}$, formulated as:
\begin{equation}\label{eq:sf}
\small
h_i=\alpha W_{S} x_i +\sum_{j \in \mathcal{N}(i)} \alpha_{ij} W_{S} x_i
\end{equation}
where each $x_i$ is a member of $X$ and $W_{S}$ is a model parameter for this layer, performing layer multiplications.
$\alpha$ can be computed by:
\begin{equation}
\small
\alpha=\frac{\exp \left({\mathcal{A}}^{\top} \operatorname{LeakyReLU}\left(W_{S}\left[{x_{lt}} \| {x_{lt}}\right]\right)\right)}{\sum_{k \in \mathcal{N}(i) \cup\{i\}} \exp \left({\mathcal{A}}^{\top} \operatorname{LeakyReLU}\left(W_{S}\left[x_{lt} \| {x_{lt}}\right]\right)\right)}
\end{equation}

\noindent \textbf{Final Knowledge Graph.}
Finally, we proceed to eliminate redundant edges by leveraging the enhanced semantic feature similarity within the node features set $H$, which comprises $h_1, h_2, \ldots, h_N$. Following this, we systematically remove all edges that fall below a predefined threshold value $\gamma$. This process results in our final KG, denoted as $\mathcal{G}_f = (\mathcal{V}, \mathcal{E}_f, H)$. An adjacency matrix $\mathcal{A}_f$ of dimensions $N \times N$ can be derived from the final graph $\mathcal{G}_f$.

\begin{table}[t]
\footnotesize
\centering
\begin{tabular}{| c | c c | c c | }
\toprule
\textbf{Text} & \textbf{SF} & \textbf{A-SFAS}  & \textbf{FDN} & \textbf{A-SFAS}\\
\midrule
\multirow{2}{*}{Poem} 
& \ding{55} &  0.451 & \ding{55} &  0.528 \\
& \ding{51} & 0.892  & \ding{51} &  0.892  \\
\hline
\multirow{2}{*}{Wiki} 
& \ding{55}  &  0.527 & \ding{55} & 0.618  \\
& \ding{51} &  0.912 & \ding{51} & 0.912 \\
\bottomrule
\end{tabular}
\caption{Impact of Semantic Filtering (SF) and Feature Denoising (FDN).}
\label{tab:ablation}
\vspace{-4mm}
\end{table}

\section{Experiments}
\vspace{-2mm}
\textbf{Experimental Setup.}
In the absence of automatic KG construction methods in the Bengali language, our research focuses on case study analysis involving two different text types: poems and Wikipedia articles. Additionally, we conduct ablation studies to dissect the importance of various components within our approach, shedding light on their individual contributions. We used the Average Semantic Feature Alignment Score (A-SFAS), a cosine similarity based metric to calculate semantic similarity \cite{zhelezniak-etal-2019-correlation,8947segg433} as a metric, which quantifies how closely related the feature vectors of nodes are, providing insights into the semantic consistency of the graph. A higher A-SFAS implies that nodes in the graph have more semantically aligned features, which can indicate higher quality in terms of feature representation.

\bigskip
\noindent \textbf{Implementation Details.}
We configure BERT embeddings with a length of 728 and set the node feature-length ($F$) to 128. The value of hyperparameter $k$ in Eq. \ref{eq:local-neighbourhood-relations} is set to 3. $\gamma$ is contingent upon the specific characteristics of the graph. In our experimental framework, $\gamma$ is configured to maintain a threshold corresponding to the retention of 90\% of the edges. Our model is trained in a fully unsupervised manner \citep{liu2021auto,veira2019unsupervised,rony2022tree}. ChatGPT (3.5 version; October 1, 2023 to October 19, 2023) is used as LLM for the experiments. 
\vspace{-1mm}
\subsection{Ablation Studies}
The \textbf{A-SFAS} metric calculates the average cosine distance between the related node features to measure the semantic alignment. From Table \ref{tab:ablation}, we can see that the scores with Semantic Filtering (\textbf{SF}) are better, meaning that SF enhances the overall performance of the model by improving node features and edge relations. In addition, we also observe that the scores are better when Feature DeNoising (\textbf{FDN}) is enabled, denoting the necessity of feature alignment with the graph by denoising.


\vspace{-1mm}
\subsection{Case Study} \label{case-study}
In this section, we will explore two KGs generated by our model using different types of text. Since we lack a graph construction tool for Bengali text, we have manually created these graphs by carefully assembling nodes and edges based on the available information.

\bigskip 
\noindent \textbf{Case Study 1: Poem.}\quad  
To show the universality of our research, we initially introduce a KG constructed from a poem authored by Amaresh Biswas, accessible via this GitHub repository\footnote{\url{https://github.com/azminewasi/BanglaAutoKG/}}. This poem is written in the historical context of the Bangladesh Liberation War of 1971. Remarkably, the generated KG accurately encapsulates the overarching theme and information presented in the poem. The central node within the graph, which denotes "independence" in English, is cohesively expressed by its supporting nodes. In addition, this KG effectively conveys the profound narrative of struggle, promise, and sacrifice made by the people of Bangladesh in their struggle for independence.

\bigskip
\noindent \textbf{Case Study 2: Wikipedia Article.}\quad 
This KG is derived from the initial two paragraphs of the Bengali Wikipedia page pertaining to the Bangladesh Football Team, accessible via this URL\footnote{\url{https://bn.wikipedia.org/s/1cmz}}. The KG successfully captures various entities, dates, and names associated with the Bangladesh football team, providing a comprehensive representation of the KG. Of particular note is the inclusion of specific details about Bangladesh's first international football match, which are effectively represented in both the text and the KG. This KG serves as a robust and easily understood representation of the source text, making it highly valuable for fact-checking purposes.

\begin{figure}[t]
\begin{center}
\includegraphics[scale=0.38]{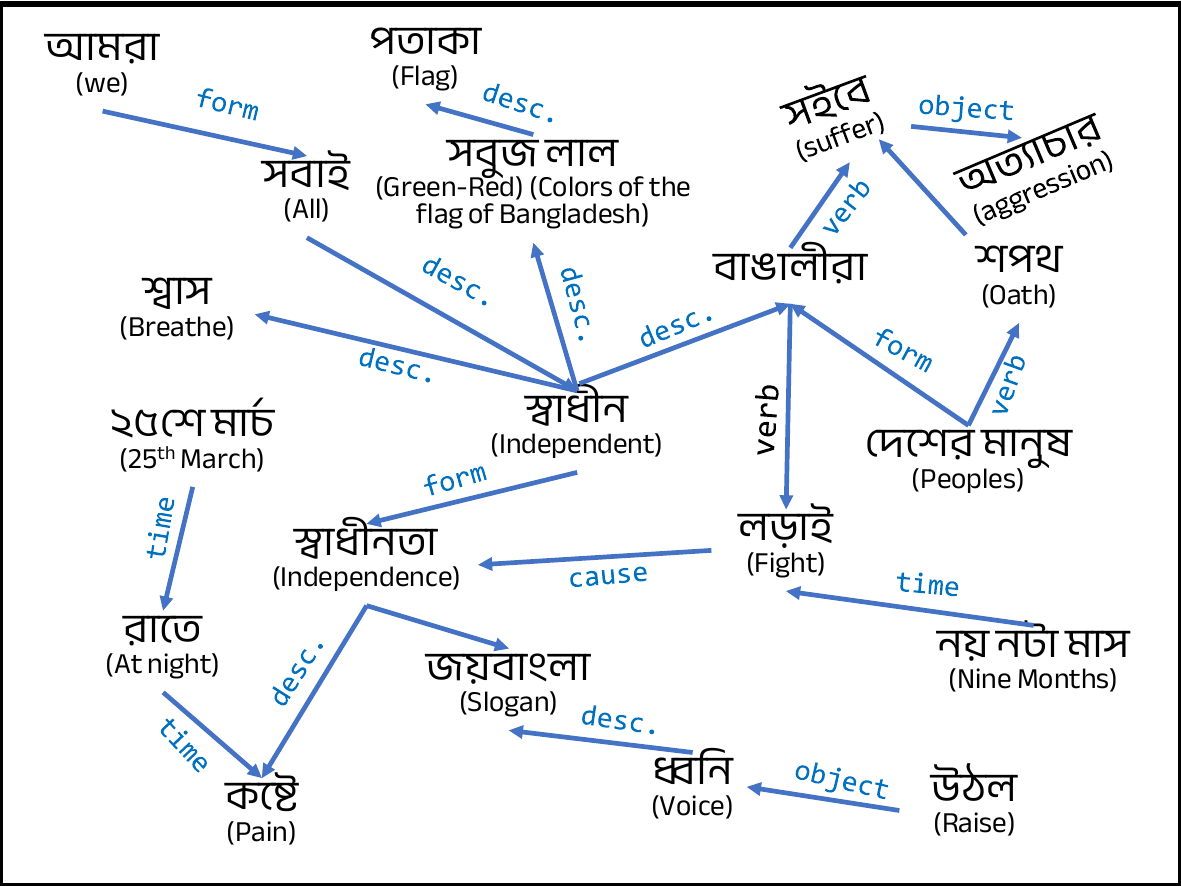} 
\vspace{-2mm}
\caption{Case Study: KG of a Poem.}
\label{fig.poem}
\vspace{-2mm}
\end{center}
\end{figure}

\begin{figure}[t]
\begin{center}
\includegraphics[scale=0.38]{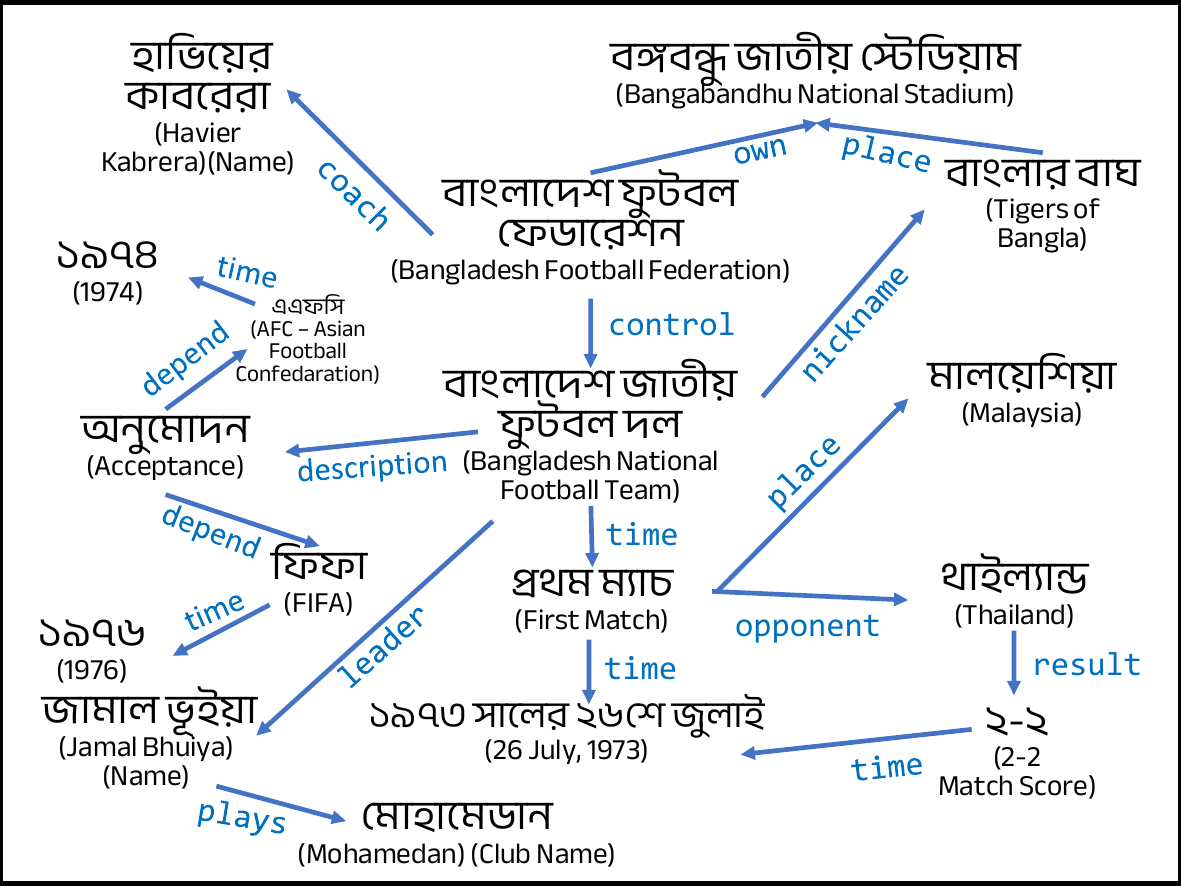} 
\vspace{-2mm}
\caption{Case Study: KG of a Wikipedia section.}
\label{fig.wiki}
\vspace{-2mm}
\end{center}
\end{figure}

\section{Discussion}
We believe it will be a starting point for working with Bangla KG problems. Though Bangla is the sixth most spoken language in the world, the growth in Bangla KG problems is almost zero. Bangla KGs can enable efficient information retrieval, knowledge discovery, fact-checking, and intelligent applications across various domains. They can contribute to preserving Bangla culture and heritage while fostering innovation and economic growth. This initiative has the potential to empower millions of Bangla speakers globally and open up a wide range of new possibilities.

The Bangla (Bengali) language is evolving rapidly, and some texts from or predating the Rabindranath Tagore \cite{DaswFfwaf2013} era may pose challenges for multilingual LLMs in terms of comprehension. Enhancements in model capabilities for such texts would undoubtedly enhance our model's overall performance, while any degradation in these aspects could impact its efficacy. Additionally, the complexity of deep metaphorical texts presents challenges for LLMs, potentially diminishing our model's performance. Another problem is irregular output from LLMs. We have also observed that when dealing with lengthy texts, the need to prompt the model multiple times to extract entities from the complete text can be cumbersome, due to short context length of some LLMs.

With enhanced understanding and generating capabilities, recently introduced LLMs such as GPT-4, Gemini, Claude 3, Mistral Large and Mixtral are built to handle large context lengths,better reasoning capabilities in a wide range of languages, including metaphorical writings. They are better able to capture the subtleties and complexities of languages thanks to their larger model size and sophisticated training methodologies. Furthermore, the problems of inconsistent outputs and the requirement for repeated reminders are lessened by their capacity to process longer contexts and produce more coherent and consistent outputs. These LLMs show promise in addressing multilingual representation, managing intricate linguistic structures, and preserving context across lengthy texts as they develop further.

Future avenues of research could involve training LLMs specifically with old Bengali texts to enhance their effectiveness on historical and archaic content. Also, the development of more robust and Bengali-friendly text encoders will significantly enhance our model's capabilities. Specifically, dedicated training on a curated corpus of ancient Bengali literature, such as works by Rabindranath Tagore, Bankimchandra Chattopadhyay, and other prominent authors \cite{dHubert2018,Chakrabarty_1991}, can help LLMs better understand and generate text in the archaic linguistic styles and metaphorical expressions prevalent in those eras. 

Studies might also focus on developing text encoders that have been tailored to handle the unique characteristics of the Bengali language, like its complicated script, varied morphology, and elaborate syntactic structures from different sources. Another crucial aspect is the creation of a large-scale, high-quality Bengali KG dataset. This dataset should contain KGs from various domains, such as history, literature, culture, and science, and can be constructed through a combination of automated extraction from reliable sources using BanglaAutoKG and human curation.

\vspace{-1mm}
\section{Conclusion}
\vspace{-2mm}
Our novel approach to developing Bangla KGs automatically leverages multilingual LLMs for entity extraction and relationship identification from diverse textual sources. By incorporating translation dictionaries and using Graph Neural Networks with self-supervised attention and semantic filtering, we automate KG construction and improve semantic accuracy. Empirical evidence and case studies demonstrate BanglaAutoKG's ability to autonomously build Bengali KGs from any given text: The case studies have offered a convincing display of our model's effectiveness on different data types; our ablation studies have highlighted the importance of different model components.
\vspace{-1mm}
\section{Acknowledgement}
\vspace{-2mm}
We extend our appreciation to Professor Ayesha Tasnim, Department of Computer Science and Engineering, Shahjalal University of Science and Technology, for her early insights and encouragement, which played an important role in shaping the initial concept of this research. 

Our work was partly supported by (1) the National Research Foundation of Korea (NRF) grant funded by the Korea government (*MSIT) (No.2018R1A5A7059549) and (2) the Institute of Information \& communications Technology Planning \& Evaluation (IITP) grant funded by the Korea government (MSIT) (No.2020-0-01373,Artificial Intelligence Graduate School Program (Hanyang University)). *Ministry of Science and ICT

\clearpage
\newpage

\section{Bibliographical References}\label{sec:reference}
\bibliographystyle{lrec-coling2024-natbib}
\bibliography{lrec-coling2024}

\begin{thebibliography}{33}
\expandafter\ifx\csname natexlab\endcsname\relax\def\natexlab#1{#1}\fi

\bibitem[{Bao et~al.(2016)Bao, Duan, Yan, Zhou, and Zhao}]{bao2016constraint}
Junwei Bao, Nan Duan, Zhao Yan, Ming Zhou, and Tiejun Zhao. 2016.
\newblock Constraint-based question answering with knowledge graph.
\newblock In \emph{Proceedings of COLING 2016, the 26th international
  conference on computational linguistics: technical papers}, pages 2503--2514.

\bibitem[{Bhattacharjee et~al.(2022)Bhattacharjee, Hasan, Ahmad, Mubasshir,
  Islam, Iqbal, Rahman, and Shahriyar}]{bhattacharjee-etal-2022-banglabert}
Abhik Bhattacharjee, Tahmid Hasan, Wasi Ahmad, Kazi~Samin Mubasshir, Md~Saiful
  Islam, Anindya Iqbal, M~Sohel Rahman, and Rifat Shahriyar. 2022.
\newblock Banglabert: Language model pretraining and benchmarks for
  low-resource language understanding evaluation in bangla.
\newblock In \emph{Findings of the Association for Computational Linguistics:
  NAACL 2022}, pages 1318--1327.

\bibitem[{Bhattacharjee et~al.(2023)Bhattacharjee, Hasan, Ahmad, and
  Shahriyar}]{bhattacharjee2023banglanlg}
Abhik Bhattacharjee, Tahmid Hasan, Wasi Ahmad, and Rifat Shahriyar. 2023.
\newblock Banglanlg and banglat5: Benchmarks and resources for evaluating
  low-resource natural language generation in bangla.
\newblock In \emph{Findings of the Association for Computational Linguistics:
  EACL 2023}, pages 714--723.

\bibitem[{Chakrabarty(1991)}]{Chakrabarty_1991}
Dipesh Chakrabarty. 1991.
\newblock \href {https://doi.org/10.2307/2057629} {Europe reconsidered:
  Perceptions of the west in nineteenth century bengal. by tapan raychaudhuri.
  delhi: Oxford university press, 1988. xviii, 369 pp. rs. 170.00.}
\newblock \emph{The Journal of Asian Studies}, 50(3):723–724.

\bibitem[{Das et~al.(2013)Das, Basu, and Mitra}]{DaswFfwaf2013}
Suprabhat Das, Anupam Basu, and Pabitra Mitra. 2013.
\newblock \href {https://doi.org/10.4018/978-1-4666-3970-6.ch013} {\emph{The
  Bengali Literary Collection of Rabindranath Tagore: Search and Study of
  Lexical Richness}}, page 302–314. IGI Global.

\bibitem[{Defferrard et~al.(2016)Defferrard, Bresson, and
  Vandergheynst}]{cnn_graph}
Micha\"el Defferrard, Xavier Bresson, and Pierre Vandergheynst. 2016.
\newblock Convolutional neural networks on graphs with fast localized spectral
  filtering.
\newblock In \emph{Advances in Neural Information Processing Systems}.

\bibitem[{Devlin et~al.(2019)Devlin, Chang, Lee, and
  Toutanova}]{DBLP:conf/naacl/DevlinCLT19}
Jacob Devlin, Ming{-}Wei Chang, Kenton Lee, and Kristina Toutanova. 2019.
\newblock \href {https://doi.org/10.18653/V1/N19-1423} {{BERT:} pre-training of
  deep bidirectional transformers for language understanding}.
\newblock In \emph{Proceedings of the 2019 Conference of the North American
  Chapter of the Association for Computational Linguistics: Human Language
  Technologies, {NAACL-HLT} 2019, Minneapolis, MN, USA, June 2-7, 2019, Volume
  1 (Long and Short Papers)}, pages 4171--4186. Association for Computational
  Linguistics.

\bibitem[{d’Hubert(2018)}]{dHubert2018}
Thibaut d’Hubert. 2018.
\newblock \href {https://doi.org/10.1093/acrefore/9780190277727.013.39}
  {Literary history of bengal, 8th-19th century ad}.
\newblock \emph{Oxford Research Encyclopedia of Asian History}.

\bibitem[{Ekram et~al.(2022)Ekram, Rahman, Altaf, Islam, Rahman, Rahman,
  Hossain, and Kamal}]{ekram2022banglarqa}
Syed Mohammed~Sartaj Ekram, Adham~Arik Rahman, Md~Sajid Altaf, Mohammed~Saidul
  Islam, Mehrab~Mustafy Rahman, Md~Mezbaur Rahman, Md~Azam Hossain, and Abu
  Raihan~Mostofa Kamal. 2022.
\newblock Banglarqa: A benchmark dataset for under-resourced bangla language
  reading comprehension-based question answering with diverse question-answer
  types.
\newblock In \emph{Findings of the Association for Computational Linguistics:
  EMNLP 2022}, pages 2518--2532.

\bibitem[{Ganea and Hofmann(2017)}]{ganea2017deep}
Octavian-Eugen Ganea and Thomas Hofmann. 2017.
\newblock Deep joint entity disambiguation with local neural attention.
\newblock In \emph{Proceedings of the 2017 Conference on Empirical Methods in
  Natural Language Processing}, pages 2619--2629.

\bibitem[{Guo et~al.(2015)Guo, Wang, Wang, Wang, and Guo}]{guo2015semantically}
Shu Guo, Quan Wang, Bin Wang, Lihong Wang, and Li~Guo. 2015.
\newblock Semantically smooth knowledge graph embedding.
\newblock In \emph{Proceedings of the 53rd Annual Meeting of the Association
  for Computational Linguistics and the 7th International Joint Conference on
  Natural Language Processing (Volume 1: Long Papers)}, pages 84--94.

\bibitem[{Islam et~al.(2021)Islam, Kar, Islam, and Amin}]{islam2021sentnob}
Khondoker~Ittehadul Islam, Sudipta Kar, Md~Saiful Islam, and Mohammad~Ruhul
  Amin. 2021.
\newblock Sentnob: A dataset for analysing sentiment on noisy bangla texts.
\newblock In \emph{Findings of the Association for Computational Linguistics:
  EMNLP 2021}, pages 3265--3271.

\bibitem[{Khorashadizadeh et~al.(2023)Khorashadizadeh, Mihindukulasooriya,
  Tiwari, Groppe, and Groppe}]{khorashadizadeh2023exploring}
Hanieh Khorashadizadeh, Nandana Mihindukulasooriya, Sanju Tiwari, Jinghua
  Groppe, and Sven Groppe. 2023.
\newblock Exploring in-context learning capabilities of foundation models for
  generating knowledge graphs from text.
\newblock In \emph{TEXT2KG/BiKE@ESWC}.

\bibitem[{Kim and Oh(2021)}]{kim2021howwfwf}
Dongkwan Kim and Alice Oh. 2021.
\newblock \href {https://openreview.net/forum?id=Wi5KUNlqWty} {How to find your
  friendly neighborhood: Graph attention design with self-supervision}.
\newblock In \emph{International Conference on Learning Representations}.

\bibitem[{Le and Titov(2018)}]{le2018improving}
Phong Le and Ivan Titov. 2018.
\newblock \href {https://doi.org/10.18653/v1/P18-1148} {Improving entity
  linking by modeling latent relations between mentions}.
\newblock In \emph{Proceedings of the 56th Annual Meeting of the Association
  for Computational Linguistics (Volume 1: Long Papers)}, pages 1595--1604.

\bibitem[{Liu et~al.(2021)Liu, You, Wu, Ge, Sun et~al.}]{liu2021auto}
Fenglin Liu, Chenyu You, Xian Wu, Shen Ge, Xu~Sun, et~al. 2021.
\newblock Auto-encoding knowledge graph for unsupervised medical report
  generation.
\newblock \emph{Advances in Neural Information Processing Systems},
  34:16266--16279.

\bibitem[{Morris et~al.(2019)Morris, Ritzert, Fey, Hamilton, Lenssen, Rattan,
  and Grohe}]{10.1609/aaai.v33i01.33014602}
Christopher Morris, Martin Ritzert, Matthias Fey, William~L Hamilton, Jan~Eric
  Lenssen, Gaurav Rattan, and Martin Grohe. 2019.
\newblock Weisfeiler and leman go neural: Higher-order graph neural networks.
\newblock In \emph{Proceedings of the AAAI conference on artificial
  intelligence}, volume~33, pages 4602--4609.

\bibitem[{Park et~al.(2022)Park, Chae, Bae, Park, and Kim}]{sakg}
Sung-Jun Park, Dong-Kyu Chae, Hong-Kyun Bae, Sumin Park, and Sang-Wook Kim.
  2022.
\newblock Reinforcement learning over sentiment-augmented knowledge graphs
  towards accurate and explainable recommendation.
\newblock In \emph{Proceedings of the Fifteenth ACM International Conference on
  Web Search and Data Mining}, pages 784--793.

\bibitem[{Peng et~al.(2019)Peng, Huh, Ling, and
  Banko}]{peng-etal-2019-improving}
Boya Peng, Yejin Huh, Xiao Ling, and Michele Banko. 2019.
\newblock Improving knowledge base construction from robust infobox extraction.
\newblock In \emph{Proceedings of the 2019 Conference of the North American
  Chapter of the Association for Computational Linguistics: Human Language
  Technologies, Volume 2 (Industry Papers)}, pages 138--148.

\bibitem[{Ren et~al.(2016)Ren, He, Qu, Voss, Ji, and Han}]{ren2016label}
Xiang Ren, Wenqi He, Meng Qu, Clare~R Voss, Heng Ji, and Jiawei Han. 2016.
\newblock Label noise reduction in entity typing by heterogeneous partial-label
  embedding.
\newblock In \emph{Proceedings of the 22nd ACM SIGKDD international conference
  on Knowledge discovery and data mining}, pages 1825--1834.

\bibitem[{Rony et~al.(2022)Rony, Chaudhuri, Usbeck, and Lehmann}]{rony2022tree}
Md~Rashad Al~Hasan Rony, Debanjan Chaudhuri, Ricardo Usbeck, and Jens Lehmann.
  2022.
\newblock Tree-kgqa: an unsupervised approach for question answering over
  knowledge graphs.
\newblock \emph{IEEE Access}, 10:50467--50478.

\bibitem[{Sazzed(2022)}]{sazzed-2022-annotated}
Salim Sazzed. 2022.
\newblock An annotated dataset and automatic approaches for discourse mode
  identification in low-resource bengali language.
\newblock In \emph{Proceedings of the Workshop on Multilingual Information
  Access (MIA)}, pages 9--15.

\bibitem[{Shang et~al.(2020)Shang, Zhang, Liu, Li, and Han}]{shang2020nettaxo}
Jingbo Shang, Xinyang Zhang, Liyuan Liu, Sha Li, and Jiawei Han. 2020.
\newblock Nettaxo: Automated topic taxonomy construction from text-rich
  network.
\newblock In \emph{Proceedings of the Web Conference 2020}, pages 1908--1919.

\bibitem[{Sitikhu et~al.(2019)Sitikhu, Pahi, Thapa, and Shakya}]{8947segg433}
Pinky Sitikhu, Kritish Pahi, Pujan Thapa, and Subarna Shakya. 2019.
\newblock A comparison of semantic similarity methods for maximum human
  interpretability.
\newblock In \emph{2019 Artificial Intelligence for Transforming Business and
  Society (AITB)}, volume~1, pages 1--4.

\bibitem[{Veira et~al.(2019)Veira, Keng, Padmanabhan, and
  Veneris}]{veira2019unsupervised}
Neil Veira, Brian Keng, Kanchana Padmanabhan, and Andreas~G Veneris. 2019.
\newblock Unsupervised embedding enhancements of knowledge graphs using textual
  associations.
\newblock In \emph{IJCAI}, pages 5218--5225.

\bibitem[{Veličković et~al.(2018)Veličković, Cucurull, Casanova, Romero,
  Liò, and Bengio}]{veličković2018graph}
Petar Veličković, Guillem Cucurull, Arantxa Casanova, Adriana Romero, Pietro
  Liò, and Yoshua Bengio. 2018.
\newblock \href {https://openreview.net/forum?id=rJXMpikCZ} {Graph attention
  networks}.
\newblock In \emph{International Conference on Learning Representations}.

\bibitem[{Wang et~al.(2018)Wang, Zhang, Wang, Zhao, Li, Xie, and
  Guo}]{wang2018ripplenet}
Hongwei Wang, Fuzheng Zhang, Jialin Wang, Miao Zhao, Wenjie Li, Xing Xie, and
  Minyi Guo. 2018.
\newblock Ripplenet: Propagating user preferences on the knowledge graph for
  recommender systems.
\newblock In \emph{Proceedings of the 27th ACM international conference on
  information and knowledge management}, pages 417--426.

\bibitem[{Xiong et~al.(2017)Xiong, Power, and Callan}]{xiong2017explicit}
Chenyan Xiong, Russell Power, and Jamie Callan. 2017.
\newblock Explicit semantic ranking for academic search via knowledge graph
  embedding.
\newblock In \emph{Proceedings of the 26th international conference on world
  wide web}, pages 1271--1279.

\bibitem[{Xu and Barbosa(2018)}]{xu2018neural}
Peng Xu and Denilson Barbosa. 2018.
\newblock \href {https://doi.org/10.18653/v1/N18-1002} {Neural fine-grained
  entity type classification with hierarchy-aware loss}.
\newblock In \emph{Proceedings of the 2018 Conference of the North {A}merican
  Chapter of the Association for Computational Linguistics: Human Language
  Technologies, Volume 1 (Long Papers)}, pages 16--25.

\bibitem[{Zeng et~al.(2015)Zeng, Liu, Chen, and Zhao}]{zeng2015distant}
Daojian Zeng, Kang Liu, Yubo Chen, and Jun Zhao. 2015.
\newblock Distant supervision for relation extraction via piecewise
  convolutional neural networks.
\newblock In \emph{Proceedings of the 2015 conference on empirical methods in
  natural language processing}, pages 1753--1762.

\bibitem[{Zhang et~al.(2023)Zhang, Peñuela, and Simperl}]{eswtGEWSAGREWAG}
Bohui Zhang, Albert~Meroño Peñuela, and Elena Simperl. 2023.
\newblock \href {https://doi.org/10.3233/faia230091} {Towards explainable
  automatic knowledge graph construction with human-in-the-loop}.
\newblock \emph{Frontiers in artificial intelligence and applications}.

\bibitem[{Zhelezniak et~al.(2019)Zhelezniak, Savkov, Shen, and
  Hammerla}]{zhelezniak-etal-2019-correlation}
Vitalii Zhelezniak, Aleksandar Savkov, April Shen, and Nils Hammerla. 2019.
\newblock \href {https://doi.org/10.18653/v1/N19-1100} {Correlation
  coefficients and semantic textual similarity}.
\newblock In \emph{Proceedings of the 2019 Conference of the North {A}merican
  Chapter of the Association for Computational Linguistics: Human Language
  Technologies, Volume 1 (Long and Short Papers)}, pages 951--962.

\bibitem[{Zhou et~al.(2016)Zhou, Shi, Tian, Qi, Li, Hao, and
  Xu}]{zhou2016attention}
Peng Zhou, Wei Shi, Jun Tian, Zhenyu Qi, Bingchen Li, Hongwei Hao, and Bo~Xu.
  2016.
\newblock Attention-based bidirectional long short-term memory networks for
  relation classification.
\newblock In \emph{Proceedings of the 54th annual meeting of the association
  for computational linguistics (volume 2: Short papers)}, pages 207--212.

\end{thebibliography}


\end{document}